\documentclass{article}

\usepackage{microtype}
\usepackage{graphicx}
\usepackage{subfigure}
\usepackage{booktabs} 

\usepackage[accepted]{icml2019}

\icmltitlerunning{Decomposing feature-level variation with Covariate-GPLVMs}

\usepackage[utf8]{inputenc} 
\usepackage[T1]{fontenc}    
\usepackage{url}            
\usepackage{booktabs}       
\usepackage{amsfonts}       
\usepackage{nicefrac}       
\usepackage{microtype}      

\usepackage{amsmath}
\usepackage{enumitem}
\setlist{nolistsep}

\usepackage[bf]{caption}
\usepackage{float}
\usepackage[english]{babel}
\usepackage[utf8]{inputenc}
\usepackage{todonotes}
\usepackage{tikz}
\usetikzlibrary{bayesnet}
\usepackage[titletoc, title, toc]{appendix}
\usepackage{microtype}
\usepackage[export]{adjustbox}
\usepackage[scaled=0.85]{beramono}
\usepackage{stmaryrd}

\usepackage{color}
\definecolor{mydarkblue}{rgb}{0,0.08,0.45}
\usepackage[colorlinks=true,allcolors=mydarkblue]{hyperref}

\newcommand{\N}{\mathcal{N}}
\newcommand{\TN}{\mathcal{TN}}
\newcommand{\KL}{\text{KL}}
\newcommand{\boldY}{\mathbf{Y}}
\newcommand{\boldK}{\mathbf{K}}

\newcommand{\boldI}{\mathbf{I}}

\newcommand{\boldF}{\mathbf{F}}
\newcommand{\boldf}{\mathbf{f}}
\newcommand{\boldy}{\mathbf{y}}
\newcommand{\boldx}{\mathbf{x}}
\newcommand{\boldX}{\mathbf{X}}

\newcommand{\boldz}{\mathbf{z}}
\newcommand{\boldZ}{\mathbf{Z}}

\newcommand{\GP}{\mathcal{GP}}

\newcommand{\boldzero}{\mathbf{0}}

\newcommand{\addintkernel}{\texttt{\textbf{add+int}} }
\newcommand{\intkernel}{\texttt{\textbf{int}} }
\newcommand{\addkernel}{\texttt{\textbf{add}} }

\newcommand{\Var}{\mathrm{Var}}

\usepackage{xcolor}

\begin{document}

\twocolumn[
\icmltitle{Decomposing feature-level variation with \\ Covariate Gaussian Process Latent Variable Models}




\begin{icmlauthorlist}
\icmlauthor{Kaspar Märtens}{oxstats}
\icmlauthor{Kieran R Campbell}{ubc,bccrc,ubcdsi}
\icmlauthor{Christopher Yau}{ati,uob}
\end{icmlauthorlist}

\icmlaffiliation{oxstats}{Department of Statistics, University of Oxford, UK}
\icmlaffiliation{ubc}{Department of Statistics, University of British Columbia, Canada}
\icmlaffiliation{bccrc}{BC Cancer Agency, Canada}
\icmlaffiliation{ubcdsi}{UBC Data Science Institute, Canada}
\icmlaffiliation{ati}{The Alan Turing Institute, UK}
\icmlaffiliation{uob}{Institute of Cancer and Genomic Sciences, University of Birmingham, UK}
\icmlcorrespondingauthor{Christopher Yau}{c.yau@bham.ac.uk}

\icmlkeywords{Gaussian Process, Latent Variable Model}

\vskip 0.3in
]



\printAffiliationsAndNotice{}  

\begin{abstract}
The interpretation of complex high-dimensional data typically requires the use of dimensionality reduction techniques to extract explanatory low-dimensional representations. However, in many real-world problems these representations may not be sufficient to aid interpretation on their own, and it would be desirable to interpret the model in terms of the original features themselves. Our goal is to characterise how feature-level variation depends on latent low-dimensional representations, external covariates, and non-linear interactions between the two. In this paper, we propose to achieve this through a structured kernel decomposition in a hybrid Gaussian Process model which we call the Covariate Gaussian Process Latent Variable Model (c-GPLVM). We demonstrate the utility of our model on simulated examples and applications in disease progression modelling from high-dimensional gene expression data in the presence of additional phenotypes. In each setting we show how the c-GPLVM can extract low-dimensional structures from high-dimensional data sets whilst allowing a breakdown of feature-level variability that is not present in other commonly used dimensionality reduction approaches.
\end{abstract}

\setlength{\abovedisplayskip}{3pt}
\setlength{\belowdisplayskip}{3pt}
\setlength{\abovedisplayshortskip}{3pt}
\setlength{\belowdisplayshortskip}{3pt}

\section{Introduction}

The identification of low-dimensional structure is crucial to gaining insight from complex high-dimensional data. In probabilistic models, this task can be formulated as finding a low-dimensional latent variable $\boldz_n \in \mathbb{R}^Q$ for each data point $n = 1, \ldots, N$ and a set of mappings $f^{(j)}: \boldz \mapsto \boldy^{(j)}$ for every feature $\boldy^{(j)}$, $j \in \{1, \ldots, P\}$ so that $Q \ll P$. Obtaining an informative one- or two-dimensional representation of the data is often particularly desirable, allowing us to visually interpret the patterns and relationships present in the data. 

There are a number of approaches for defining the mapping functions, ranging from linear models such as the pPCA \citep{tipping1999probabilistic} to non-linear models provided by neural networks such as the Variational Autoencoder (VAE) \citep{kingma2014auto}. The former are classically well-studied, their behaviours widely explored and can be considered to be ``interpretable'', but the linearity assumptions can often be too restrictive. Non-linear techniques offer more flexibility with the disadvantage that the non-trivial dependencies that are learnt can be challenging to characterise, leading to their popular description as ``black box techniques''. 

While the derivation of low-dimensional representations represents a powerful means of information extraction from high-dimensional data, they can be of limited utility without an explicit reference back to the original observed features. For example, in transcriptomics, latent representations of gene expression profiles can help to identify sub-populations of biological samples with similar coordinated gene behaviour. However, as the underlying biology is ultimately physically driven by variation at the level of individual genes, we would like to decompose that expression variability into a number of meaningful sub-components, in particular, as a function of latent coordinates and other observed covariates.

\begin{figure}[!t]
\centering
\includegraphics[width=\columnwidth]{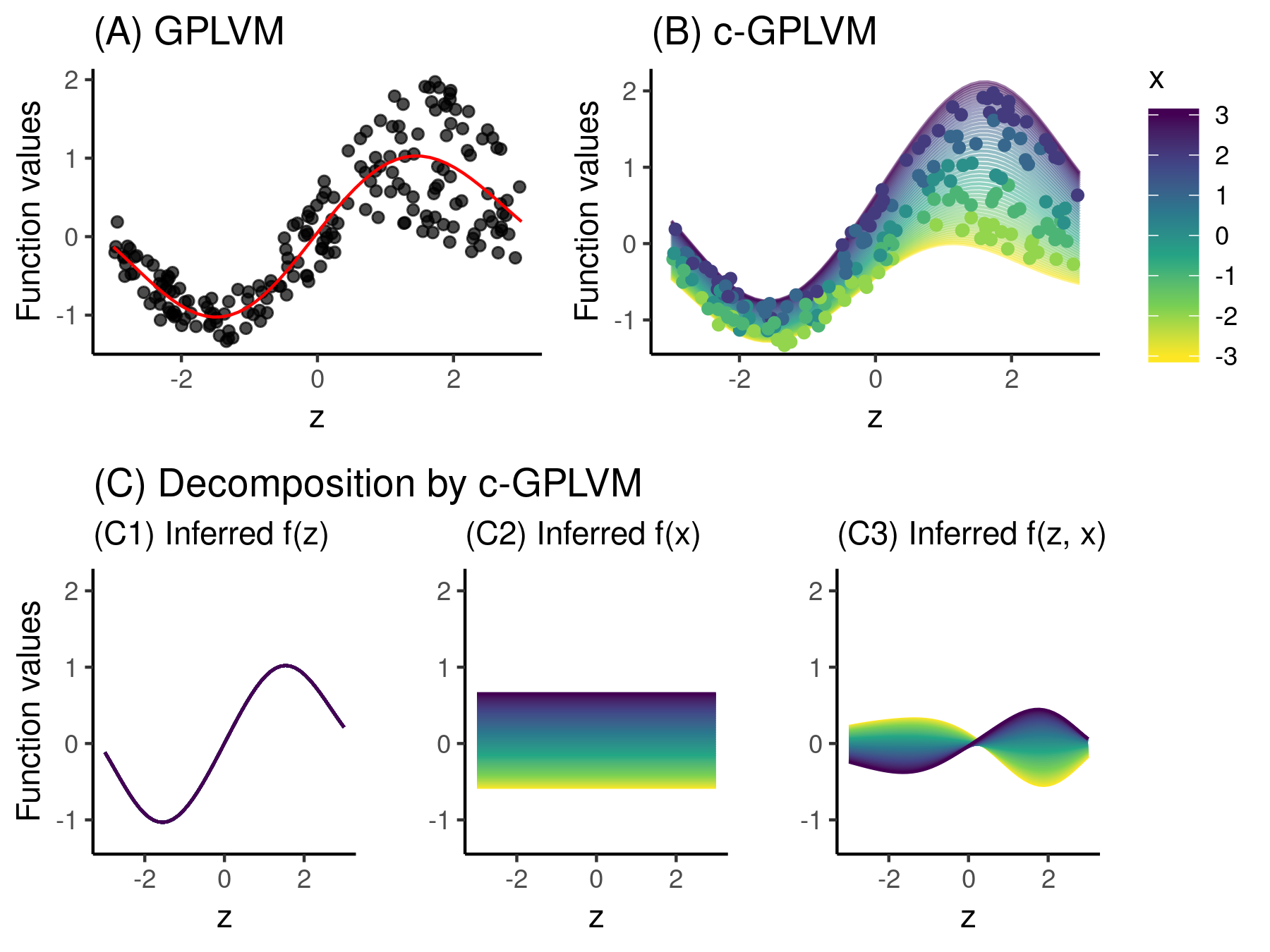}
\caption{
The presence of covariate effects confounds the mappings from the underlying latent $\boldz$ (on $x$-axis) to the feature space (selected feature on $y$-axis). Standard GPLVM (panel A)) ignores covariate information which is denoted by colour, whereas its inclusion in c-GPLVM (panel B)) allows us to capture a variety of nonlinear covariate effects. Furthermore, we would like to decompose the feature-level variation into covariate $\boldx$ and latent $\boldz$ additive contributions and their non-linear interaction as shown in panels (C1-C3). 
}
\label{fig:motivation}
\vspace{-5mm}
\end{figure}

In this paper, we focus on dimensionality reduction in the presence of additional covariate information. This is often available in real-life applications, 
for example, in transcriptomics, covariate information might include categorical labels (e.g. denoting known disease sub-populations), continuous-valued measurements (e.g. biomarkers), or censored information (e.g. patient survival times). More formally, suppose that in addition to $(\boldz_n, \boldy_n)$, each data point is also associated with a $C$-dimensional covariate vector, $\boldx_n \in \mathbb{R}^{C}$ which may act to modulate the variation in the data. 
Our goal is not black-box predictive modelling, i.e.\ we are not interested in simply learning conditional distributions $p(\boldy|\boldx)$ or $p(\boldx|\boldy)$. Instead, our work focuses on discovery applications.
Our goal is two-fold:
\begin{itemize}
    \item Learn a low-dimensional $\boldz$ that is covariate-adjusted. 
    \item Characterise feature-level variation, separating what is explained by $\boldz$ from the contribution of $\boldx$. 
\end{itemize}
Thus we are interested in inferring both the posterior $p(\boldz | \boldy, \boldx)$ over latent coordinates as well as the mappings $p(\boldf^{(j)} | \boldy, \boldx, \boldz)$ for every feature $j$. We wish to understand the nature of the feature-level variability and e.g.\ allow us to identify sets of features which are fully explained by covariates versus those which show complex dependence on both latent variables and covariates. We will principally do this in a high-dimensional data setting where the number of features is vastly greater than the number of covariates. 


We particularly focus on the use of Gaussian Processes (GP) as a non-parametric model over the mapping functions. Our choice reflects the strong theoretical underpinnings of GP models as well as recent advances that have enabled such models to be scalable to large data sets \citep{hensman2013gaussian,hensman2015classification}. In the context of dimensionality reduction, the Gaussian Process Latent Variable Model (GPLVM) \citep{lawrence2005probabilistic} is the reference for probabilistic non-linear dimensionality reduction which has spawned numerous extensions, e.g. \cite{shon2006learning,ek2009shared,gao2011supervised,jiang2012supervised,damianou2013deep,gadd2018pseudo}. 

\begin{figure}[!t]
\centering
\includegraphics[width=0.95\columnwidth]{fig/cgplvm_graphical_model.png}
\caption{Graphical models for (a) GPLVM, (b) a particular implementation of supervised-GPLVM, and (c) c-GPLVM.}
\label{fig:graphical_model}
\includegraphics[width=\columnwidth]{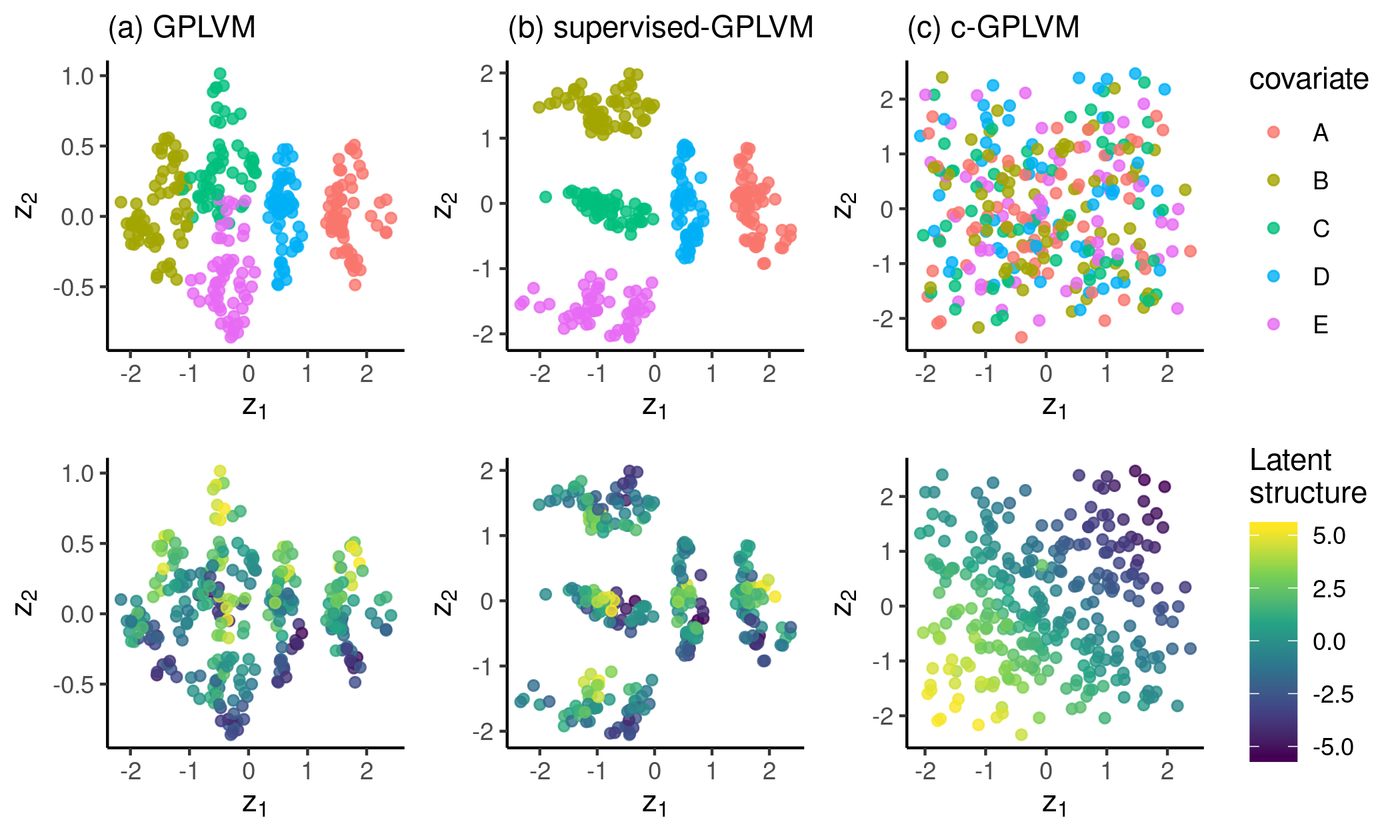}
\caption{Here we use GPLVM, supervised-GPLVM and c-GPLVM to learn a 2D latent space ($\boldz$) from a synthetic data set which contains 5 discrete classes A-E ($\boldx$). 
Here the data generative mechanism contains both the class effects (covariate $\boldx$) as well as the shared 2D latent structure ($\boldz$).
For all three methods, their learned 2D latent space $\boldz$ has been shown twice, coloured by $\boldx$ (top panel) and the true underlying $\boldz$ (bottom panel).
The dominant effect of the 5 classes means a GPLVM will learn a latent space that reflects the presence of these groups. This is further exaggerated with the supervised-GPLVM which acts to increase the separation between the classes and is most useful if class discrimination is the ultimate objective. In contrast, the c-GPLVM seeks to find the common shared structure between the 5 classes, and infers a latent space which \emph{adjusts} for the presence of the 5 classes (in linear modelling this process would be analogous to ``regressing out'' the class-specific effects). 
}
\label{fig:latentspace}
\vspace{-5mm}
\end{figure}

Figure~\ref{fig:motivation}(A,B) illustrates our setting of interest where feature values vary not only over the latent coordinate, but also over a single, continuous covariate. This dependence on the covariate may confound the mappings from $\boldz$ to $\boldy^{(j)}$ when applying a GPLVM. As a result, it will fail to account for the underlying latent structure in the data that is shared across all covariate values. Covariate effects can be incorporated in various ways (e.g.\ the supervised-GPLVM in Figure~\ref{fig:graphical_model}). 

Our proposal, which we call the \emph{covariate-GPLVM} (c-GPLVM), specifically focuses on learning a set of mappings $f^{(j)}: (\boldz, \boldx) \mapsto \boldy^{(j)}$ which are defined on the joint space of $\boldz$ and $\boldx$. This can be seen as a hybrid between the GP regression and the GP latent variable model where the input space consists of two parts: fixed covariates $\boldx$ and unobserved latent coordinates $\boldz$. Furthermore, we embed a structured sparsity-inducing kernel decomposition which allows c-GPLVM to explicitly disentangle variation in the observed data vectors induced by the covariates and/or latent variables, and the \emph{interaction} between the two (Figure~\ref{fig:motivation}C). The novelty of our approach is that the structured kernel permits both the development of a nonlinear mapping into a latent space where confounding factors are already adjusted for and feature-level variation that can be deconstructed. This construction leads to inferring a \emph{covariate-adjusted} latent space as illustrated in Figure~\ref{fig:latentspace}.

We demonstrate the utility of c-GPLVM on a number of simulated examples and applications in disease progression modelling from high-dimensional gene expression data in the presence of additional phenotypes. In each setting we show that the c-GPLVM is able to effectively extract low-dimensional structures from high-dimensional data sets whilst allowing a breakdown of feature-level variability that is not present in other commonly used dimensionality reduction approaches. 

\section{Background}

\subsection{Gaussian Process regression}

Gaussian processes offer a principled non-parametric framework for inference over functions \citep{rasmussen2006gaussian}. Consider a real valued function defined on the $D$-dimensional inputs $\boldX := (\boldx_1, \ldots, \boldx_N)$. A function $f$ is said to be drawn from a GP with mean $\boldzero$ and covariance $k(\boldx, \boldx')$, denoted by $f(\boldx) \sim \GP(\boldzero, k(\boldx, \boldx'))$, when 
\begin{align*}
p(\boldf | \boldX, \theta) = \N(\boldf | \boldzero, \boldK)
\end{align*}
where $\boldK$ is the kernel matrix of all pairs of inputs which also depends on hyperparameters $\theta$, with elements $\boldK_{ij} := k(\boldx_{i}, \boldx_{j})$. One popular choice for $k(\cdot)$ is the squared exponential ARD kernel, 
\begin{align*}
k(\boldx_{i}, \boldx_{k}) = \sigma^2 \exp \left( - \frac{1}{2} \sum_{j=1}^D \frac{(x_{ij} - x_{kj})^2}{l_j^2} \right)
\end{align*}
where $\sigma^2$ is the kernel variance parameter and $l_j$ are the feature-specific lengthscales. 
Combining the GP prior with a given likelihood $p(\boldy | \boldf)$ can lead to a variety of models, e.g. the GP regression model in case of the Gaussian likelihood.

\subsection{Gaussian Process Latent Variable Model}

GPLVM is a latent variable model which uses GPs as latent mappings, being a non-linear extension of the probabilistic PCA \citep{lawrence2005probabilistic}. 
Suppose we have an observed data matrix $\boldY$, consisting of $P$ features $(\boldy^{(1)}, \ldots, \boldy^{(P)})$ and $N$ data points, and our goal is to learn a low-dimensional representation $\boldZ := (\boldz_1, \ldots, \boldz_N)$. We place a prior over these latent variables, $p(\boldZ) = \prod_{i=1}^N \N(\boldz_i | \boldzero, \boldI)$ and our aim will be to infer the respective posterior. Now conditional on these latent inputs, the GPLVM is essentially a multi-output GP regression model as it specifies a GP prior for every latent mapping $\boldf^{(j)}$ and some likelihood $p(\boldy^{(j)} | \boldf^{(j)})$ for $j \in \{1, \ldots, P\}$. Denoting the collection of GP function values $\boldF := (\boldf^{(1)}, \ldots, \boldf^{(P)})$, the GPLVM is formulated as the following generative model
\begin{align*}
p(\boldY, \boldF, \boldZ, \theta) = p(\boldY | \boldF) p(\boldF | \boldZ, \theta) p(\boldZ)
\end{align*}
In the special case when the emission likelihood is Gaussian, i.e.\ when $p(\boldy^{(j)} | \boldf^{(j)}) = \N(\boldy^{(j)} | \boldf^{(j)}, \sigma^2 \boldI)$, one can analytically integrate out the GP mappings, resulting in the marginal likelihood
\begin{align} \label{eq:gplvm}
p(\boldY | \boldZ, \theta) = \prod_{j=1}^P \N(\boldy^{(j)} \,|\, \boldzero, \boldK_{zz}^{(j)} + \sigma_j^2 \boldI)
\end{align}
In this special case, we do not need to infer $\boldF$ explicitly and now inference needs to be carried out on latent variables $\boldz_1, \ldots, \boldz_N$ and kernel hyperparameters $\theta$ only. 



\section{Covariate-GPLVM}

We now consider our extension of the GPLVM to include fixed inputs where we are specifically interested in understanding the \emph{interaction} between the measured covariates $\boldx$ and latent variables $\boldz$.

\subsection{Model}

Specifically, we aim to learn mappings which are defined on the joint space of $\boldz$ and $\boldx$, i.e. $f^{(j)} : (\boldz, \boldx) \mapsto \boldy^{(j)}$. As discussed earlier, this is crucial for letting us to learn a \emph{covariate-adjusted} representation $\boldz$ (as opposed to for example modelling both $\boldx$ and $\boldy$ as a function of $\boldz$). 

Different assumptions about the form of interaction between $\boldz$ and $\boldx$ can be made, and these can be encoded via different kernel structures. One approach would be to define the ARD kernel on this joint space:
\begin{align*}
    k&^{\text{\intkernel}}((\boldx, \boldz), (\boldx', \boldz')) := \\
     &\sigma^2_{xz} \exp \left[ -\frac{1}{2} \sum_{j=1}^{C} \left( \frac{x_{ij} -  x_{kj}}{l_j^{(x)}} \right)^2-\frac{1}{2} \sum_{j=1}^{Q} \left( \frac{z_{ij} - z_{kj}}{l_j^{(z)}} \right)^2 \right] .
\end{align*}
Alternatively, one could define an additive kernel
\begin{align*}
k^{\text{\addkernel}}((\boldx, \boldz), (\boldx', \boldz')) := 
k^{x}(\boldx, \boldx') + k^{z}(\boldz, \boldz')
\end{align*}
where both $k^{x}$ and $k^{z}$ are ARD kernels. 

\begin{figure}
    \centering
    \includegraphics[width=0.9\columnwidth]{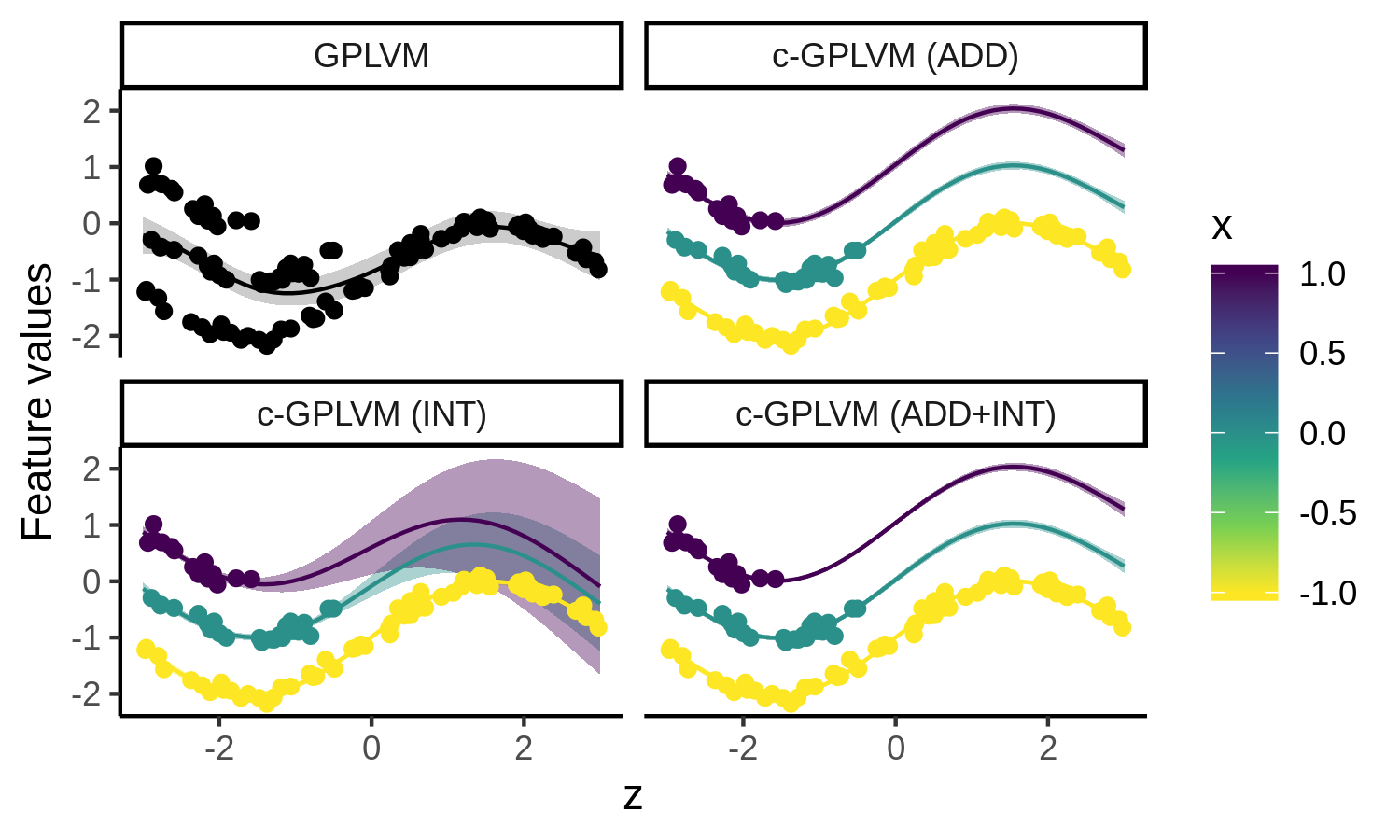}
    \caption{ADD and INT kernels have different behaviour when prediction requires extrapolation from training data. Observed data is shown as dots ($\boldz$ on $x$-axis, selected feature on $y$-axis, coloured by covariate $\boldx$). Predictions by the c-GPLVM together with uncertainty for $\boldx \in \{-1, 0, 1\}$.}
    \label{fig:add_int_extrapolation}
    \vspace{-5mm}
\end{figure}

Using either an additive (ADD) model or an interaction (INT) model correspond to different modelling assumptions. We have illustrated the implications of this on a synthetic example in Figure~\ref{fig:add_int_extrapolation}. When trained on data with covariate values $\boldx_i \in \{-1, 0, 1\}$ and the inferred $\boldz_i$ values cover the entire range from -3 to 3. However, training data with covariate $\boldx=1$ is not well represented (e.g. it could be scarce in practice), and thus prediction for future observations with $\boldx=1$ may require extrapolation. Note that even though the ADD model is more restrictive, it is more data-efficient when the feature dependence w.r.t.\ to $z$ is stationary, and can extrapolate well in scenarios when the non-stationarity of the INT model provides too much uncertainty.

Ideally we would like to use the simpler ADD model when it is appropriate, and capture everything that remains unexplained via the INT component. However, this is a challenging task. One might try to tackle this problem via model selection, similarly to how the Automatic Statistician \citep{duvenaud2013structure} carries out search for optimal combinations of kernels. But note that in the GPLVM setting, we would have to select between ADD and INT kernels for every feature, resulting in $2^P$ possible combinations which will be computationally infeasible for a high-dimensional dataset. 

Instead, we propose an ADD+INT decomposition which will not require model search. Not only will it detect the presence or absence of additive and interaction effects (resulting in behaviour as shown in Figure~\ref{fig:add_int_extrapolation} panel ADD+INT), but it will also aid interpretation via \emph{decomposing} feature-level variation into ADD and INT effects. 





\subsection{ADD+INT decomposition}

In classical statistics, the Latent Variable Regression (LVR) model could be used to infer the following \emph{linear} decomposition,  
\begin{align*}
    y_i^{(j)} = \mu_j + \sum_k \alpha_k^{(j)} z_{ik}  + \sum_l \beta_l^{(j)} x_{il} + \sum_{k,l} \gamma_{kl}^{(j)} z_{ik} x_{il} + \varepsilon_{ij}
\end{align*}
This decomposition provides insights into how much of the variation is explained by individual inputs $\boldz_k$ and $\boldx_l$, and how much is explained by their interactions. 

We are interested in learning an analogous decomposition for GPs, i.e.\ we would like to decompose the mapping $f^{(j)}$ as follows
\begin{align} \label{eq:addint}
    f^{(j)}(\boldz, \boldx) = f_0^{(j)} + f_z^{(j)}(\boldz) + f_x^{(j)}(\boldx) + f_{zx}^{(j)}(\boldz, \boldx)
\end{align}
where $f_z^{(j)}(\boldz)$ and $f_x^{(j)}(\boldx)$ would capture marginal $\boldz$ and $\boldx$ effects, and the interaction $f_{zx}^{(j)}(\boldz, \boldx)$ would capture everything that remains \emph{unexplained} by the two. 

\subsubsection{Unidentifiability}

Below we drop index $j$ to reduce notational clutter. 

A standard approach would be to place independent GP priors over $f_0, f_z, f_x, f_{zx}$. However this results in an unidentifiable decomposition. This is because $f_0, f_z, f_x$ can all be seen as bivariate functions which have positive probability under the GP prior $f_{zx} \sim \mathcal{GP}(0, k(\cdot))$. Thus the set of functions $f_0$ is a subset of functions $f_z$ (and $f_x$). Similarly the latter are a subset of functions $f_{zx}$. In other words, the supports of the four independent GP priors overlap.  

\begin{figure}
    \centering
    \includegraphics[width=0.9\columnwidth]{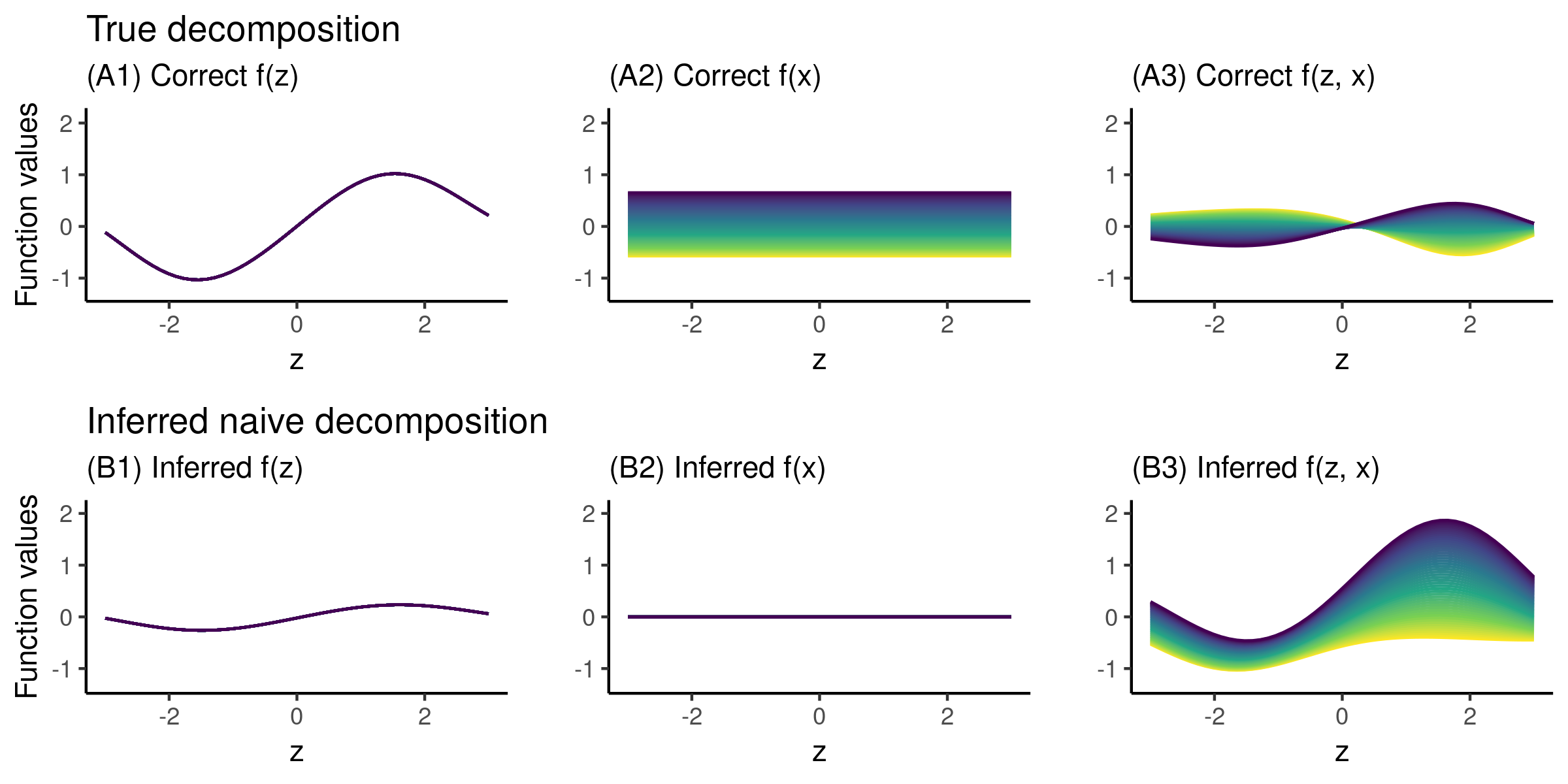}
    \caption{The naive decomposition \eqref{eq:addint} is unidentifiable. The true generative mechanism (top row) and an example of an inferred naive decomposition (bottom row).}
    \label{fig:decomposition_unidentifiable}
\end{figure}

This unidentifiability is illustrated in Figure~\ref{fig:decomposition_unidentifiable}, where on synthetic data we have shown an alternate realisation of the inferred decomposition. This alternate decomposition (in bottom row) would suggest that there is no additive $\boldx$ effect and that almost all of the variation is explained by the interaction effect, which was not the generative mechanism used to synthesise the data (shown in top row). 

\subsubsection{Zero-mean functional constraints}

To turn this into an identifiable task, we first note that for a one-dimensional $g \sim \GP(0, k())$, the distribution of $g$ conditional on $\int_a^b g(t) dt = 0$ is still a GP, but with a modified kernel \cite{durrande2011additive}. For some kernels such as the squared exponential, this can be calculated in closed form, letting us easily construct a mean-zero GP prior on some interval $[a, b]$. 

Furthermore this construction can be extended to a mean-zero interaction \cite{durrande2013anova}. To uniquely define $f_{zx}$, we need to fix its marginal distributions. A natural way would be to enforce zero-marginals $\int_a^b f_{zx}(z, x) dz = 0$ and $\int_a^b f_{zx}(z, x) dx = 0$. Similarly, we uniquely define  $f_z$ and $f_x$ by enforcing $\int_a^b f_z(z) dz = 0$, $\int_a^b f_x(x) dx = 0$. 
Now the decomposition \eqref{eq:addint} has become identifiable. 

This construction can be seen as a joint prior over the four functional subspaces $f_0, f_z, f_x, f_{zx}$ which has the following two properties: First, their supports are non-overlapping. Second, the respective functional subspaces are orthogonal in $L^2$.
The former is sufficient for a unique decomposition, but the latter makes interpretation easier, because
\begin{align*}
    \Var(f) = \Var(f_0) + \Var(f_z) + \Var(f_x) + \Var(f_{zx})
\end{align*}
due to orthogonality.

The corresponding kernel decomposition is
\begin{align*}
    k&^{\addintkernel}((\boldz, \boldx), (\boldz', \boldx')) := \\ 
    &\sigma^2_{b} + \sigma^2_z \tilde{k}(\boldz, \boldz') + \sigma^2_x \tilde{k}(\boldx, \boldx') + \sigma^2_{zx} \tilde{k}(\boldz, \boldz') \tilde{k}(\boldx, \boldx')  
\end{align*}
where $\tilde{k}(\cdot)$ is the mean-zero squared exponential kernel (details given in Supplementary). 

Finally, we make use of Bayesian shrinkage priors on the kernel variances $\sigma^2_{z}$, $\sigma^2_{x}$, $\sigma^2_{xz}$ to encourage unnecessary components to be be shrunk to zero. For this purpose, we specify $\sigma^2_{z}, \sigma^2_{x}, \sigma^2_{xz} \sim \Gamma(1, 1)$.

\subsection{Inference for c-GPLVM}

The likelihood for c-GPLVM has a similar form to \eqref{eq:gplvm}, now containing both fixed $\boldX$ and latent $\boldZ$ inputs, 
\begin{align*} \label{eq:cgplvm}
p(\boldY | \boldZ, \boldX) = \prod_{j=1}^P \N(\boldy^{(j)} | \boldzero, \boldK_{zz}^{(j)} + \boldK_{xx}^{(j)} + \boldK_{(zx)(zx)}^{(j)} + \sigma_j^2 \boldI)
\end{align*}


Variational inducing point based inference \citep{titsias2009variational, damianou2016variational} can be adopted for c-GPLVM in a straightforward way. As the kernels in c-GPLVM are defined on the extended (product) space of $\boldx$ and $\boldz$, so the inducing points now lie in this space which has dimensionality $\text{dim}(\boldx) + \text{dim}(\boldz)$. Under certain modelling assumptions this dimensionality may be reduced (e.g. under the ADD kernel we simply need inducing points in the $\boldx$ and $\boldz$ space separately and this can reduce the computational cost). 




\subsection{Inference for censored covariates}

Next, we discuss inference in the scenario when some of the inputs $\boldx$ have not been observed, but instead have been censored. In practice, this happens when we want to use patient survival times in the role of covariate, but for many individuals we only have access to their last follow-up time. We treat this as a lower bound on their true survival. In this section, we assume $\boldx_i$ are one-dimensional and represent survival times. 

Suppose for some individuals $i$ we have observed their true survival $\boldx_i$, whereas for the rest, we will treat $\boldx_i$ as latent. For the latter, we additionally have observed a lower bound $a_i$ which we want to incorporate into our model as this information is informative of the posterior of $\boldx_i$. 

Following a common assumption in survival analysis, we assume \textit{a priori} Weibull-distributed  patient survival, i.e.\ for all individuals $i$ we specify a prior $p(x_i) = \text{Weibull}(k_0, \lambda_0)$ with hyperparameters $k_0, \lambda_0$. 
Next, we incorporate censoring information by conditioning on survival $x_i$ exceeding $a_i$, and additionally upper bounding by the assumed maximum lifespan $b_i$ (if one does not want to make such an assumption, $b_i$ can be set to $\infty$). The conditional prior takes the form $p(x_i | a_i, b_i) = \text{TruncWeibull}_{[a_i, b_i]}(k_0, \lambda_0)$ which has a closed form density. 

Next we construct a variational inference scheme which will allow us to fit c-GPLVM in the presence of censored covariates as well as to infer the posteriors over true survival times. 
We need to choose the family of approximating distributions $q(\boldx_i)$ such that they would have the same support as $p(\boldx_i)$. We choose it to be the truncated Gaussian distribution which has been restricted to the interval $[a_i, b_i]$, i.e.\ we choose $q(\boldx_i) = \TN_{[a_i, b_i]}(\mu_i, \sigma_i^2)$ where $\mu_i, \sigma^2_i$ are variational parameters. In addition to having appropriate support, this choice lets us perform efficient reparameterisation-based inference \cite{kingma2014auto, rezende14}. 

This lets us sample from $q(\boldx_i)$ via reparameterisation:
\begin{align*}
    F(a_i) :&= \Phi((a_i-\mu_i)/\sigma_i) \\
    F(b_i) :&= \Phi((b_i-\mu_i)/\sigma_i) \\
    \varepsilon :&= F(a) + u (F(b)-F(a)) \text{ with } u \sim U(0, 1) \\
    x_i &\sim \Phi^{-1}(\varepsilon)\sigma_i + \mu_i
\end{align*}
where $\Phi(\cdot)$ is the cdf of the standard normal. 

Denoting the censored set of survival times by $\boldX^{\text{cens}}$ and fully observed set by $\boldX^{\text{obs}}$, we can fit c-GPLVM by optimising the following ELBO:
\begin{align*}
    \mathcal{L} = &\mathbb{E}_{q(\boldX^{\text{cens}}) q(\boldZ)} \log p(\boldY | \boldZ, \boldX^{\text{obs}}, \boldX^{\text{cens}}) + \\ 
    &- \sum_{i=1}^N \KL(q(\boldz_i) || p(\boldz_i)) 
    - \sum_{i \in \text{cens}} \KL(q(\boldx_i) || p(\boldx_i)) .
\end{align*}

\section{Related Work}

A number of alternative approaches have been developed which can be said to lie on the spectrum between GPR and GPLVMs.

Latent GP regression models (LGPR) \citep{wang2012gaussian, bodin2017latent} extend the input space of a GP regression model with additional latent variables that are used to modulate the covariance function. LGPR models often implicitly assume that the dimensionality of the latent variables is less than the number of regressors $\dim(\boldx_n) \gg \dim(\boldz_n)$, in this sense being closer to a GP regression model compared to a GPLVM. 
Approximate marginalisation of the latent variables allows these models to consider non-stationary, multimodal behaviour, and e.g.\ has recently been used for conditional density estimation~\citep{dutordoir2018gaussian}. 
In these hybrid GP models, the inputs $\boldz$ and $\boldx$ are usually concatenated, resulting in the INT kernel (as in our section 3.1) suited for predictive applications.

Semi-described and semi-supervised learning in GP regression is considered by \citep{damianou2015semi} when the inputs (or outputs) are partially observed or uncertain, this formulation also leads to a hybrid model with latent variables introduced in place of missing data.

Supervised GPLVM encompass a family of related ideas seek to decompose the joint distribution $p(\boldy,\boldx,\boldz)$ in different ways. For example, in \citep{gadd2018pseudo}, the joint distribution is factorised such that the latent variables are conditionally dependent on the covariates $p(\boldz|\boldx)$. Whilst in the supervised-GPLVM formulation \citep{gao2011supervised} and the shared-GPLVM \citep{shon2006learning} both observations and fixed inputs are conditionally dependent on the latent inputs, $p(\boldy|\boldz)p(\boldx|\boldz)$. In the discriminative-GPLVM \citep{urtasun2007discriminative}, class label information is used to maximise the discriminatory power between discrete classes in the latent space. 
This is fundamentally different from c-GPLVM which learns a \emph{covariate-adjusted} $\boldz$. 

Similar decompositions as~\eqref{eq:addint} were also discussed in \citep{duvenaud2011additive}, but as we discussed above, such a kernel decomposition is unidentifiable. This is not problematic for predictive applications they consider, but it is crucial for interpretability. 


\section{Experiments}

Our implementation of c-GPLVM is available in \url{https://github.com/kasparmartens/c-GPLVM}. 

\subsection{Toy examples}

\begin{figure}
    \centering
    \includegraphics[width=\columnwidth]{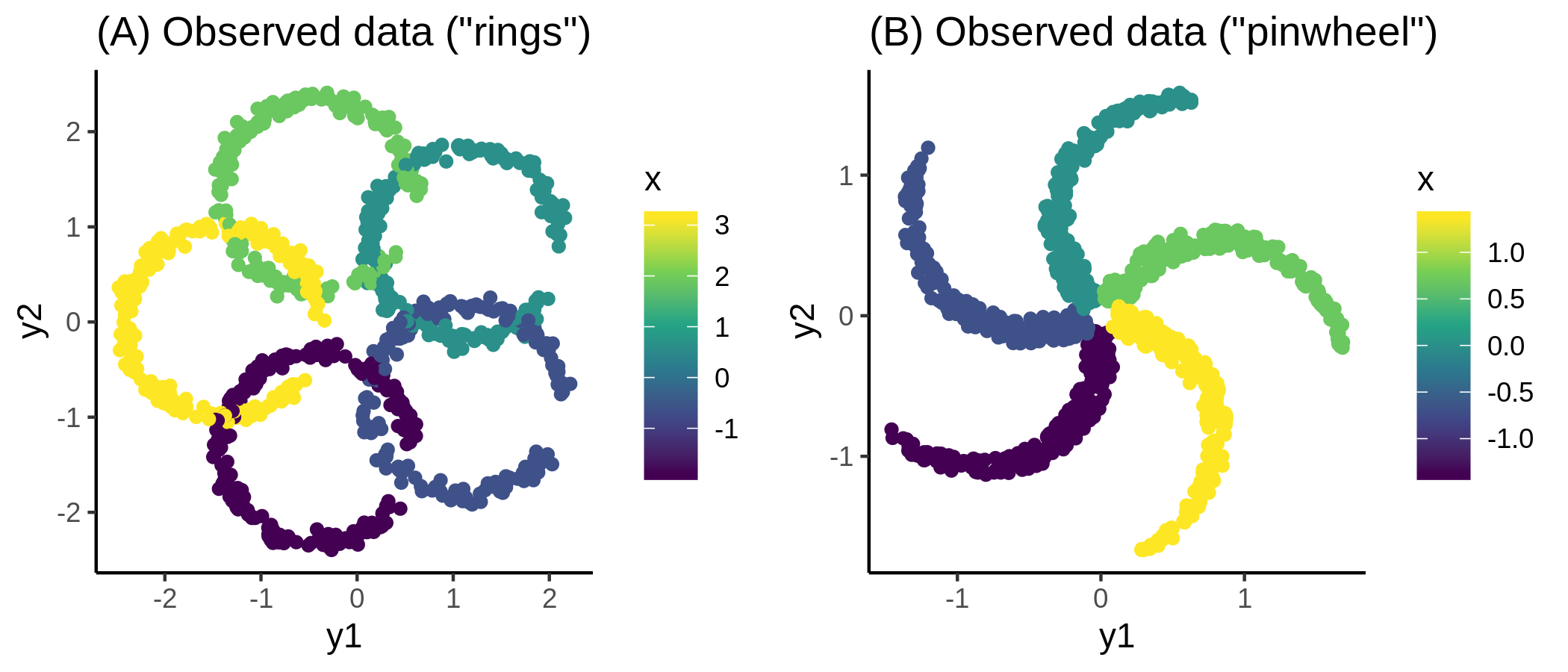}
    \caption{Observed 2D toy data with additional covariate information shown by colour. }
    \label{fig:2D_toy_data}
    \centering
    \includegraphics[width=\columnwidth]{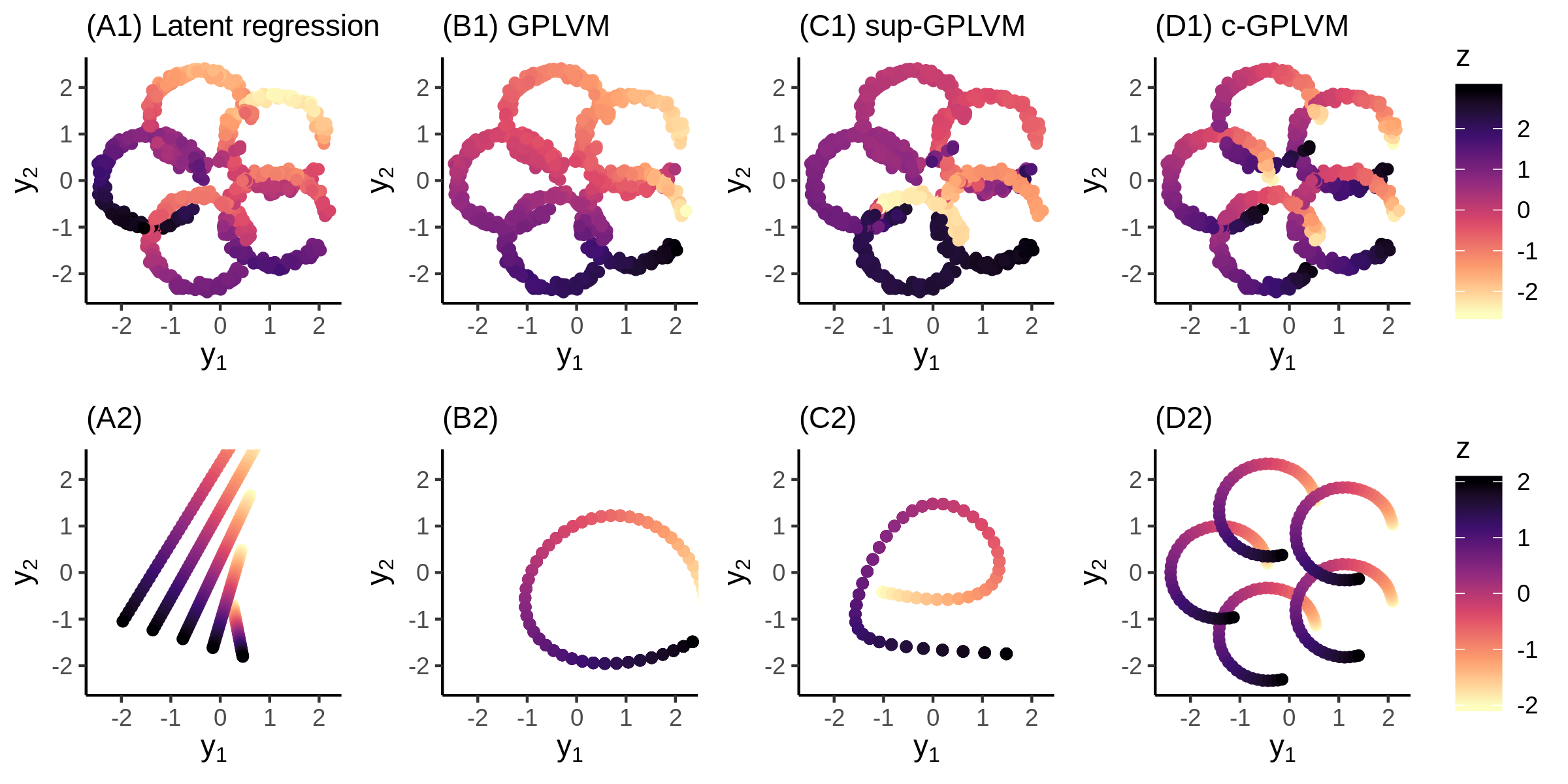}
    \caption{Comparison of models on the ``rings'' data. Top row: data coloured according to the inferred one-dimensional $\boldz$ values. Bottom row: Mean of the posterior predictive in the 2D observation space when varying $\boldz \in [-2, 2]$ and fixing $\boldx$ to the five observed values.}
    \label{fig:circles}
    \includegraphics[width=\columnwidth]{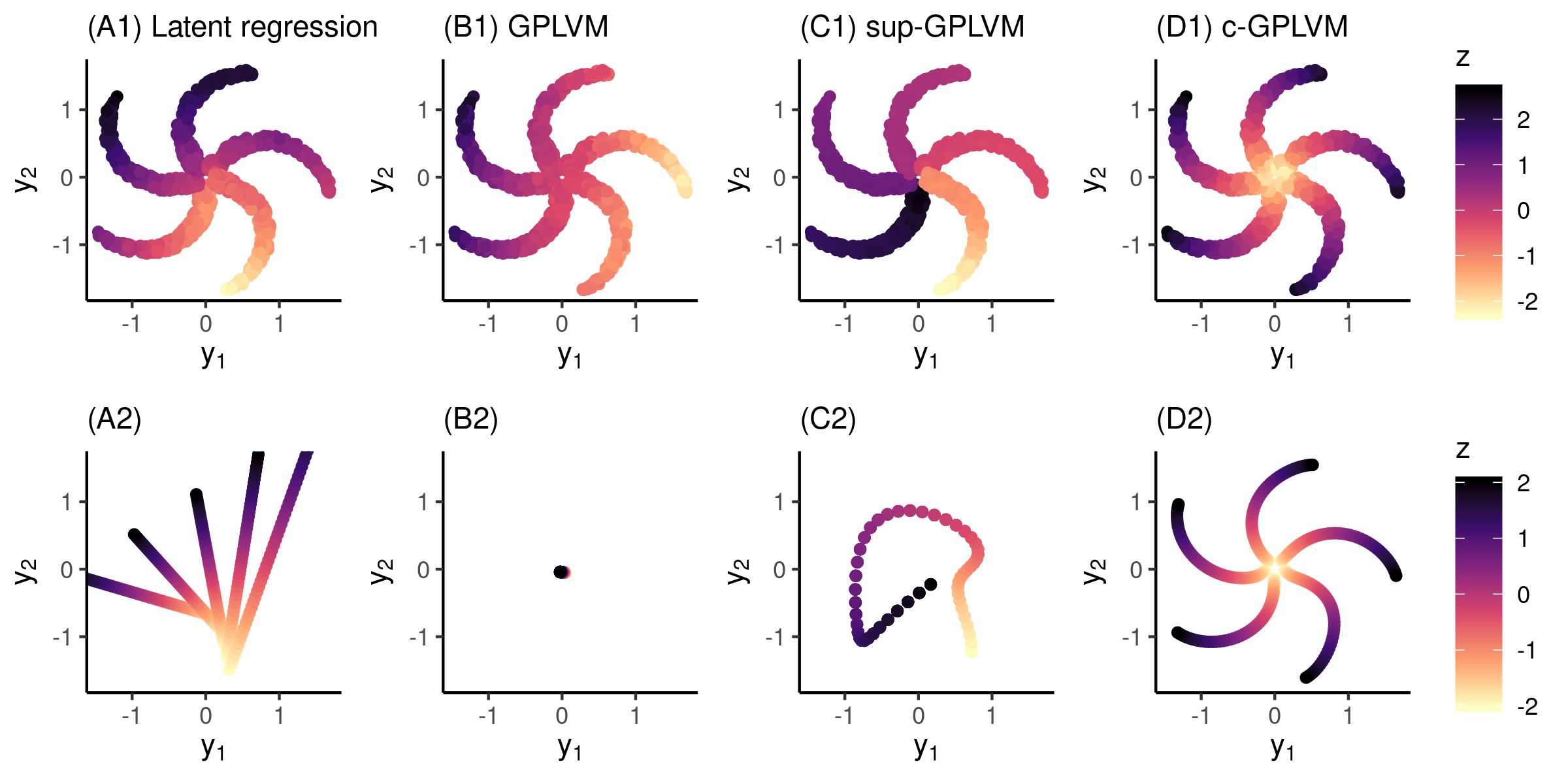}
    \caption{Comparison of models on the ``pinwheel'' data.}
    \label{fig:pinwheel}
    \centering
    \includegraphics[width=\columnwidth]{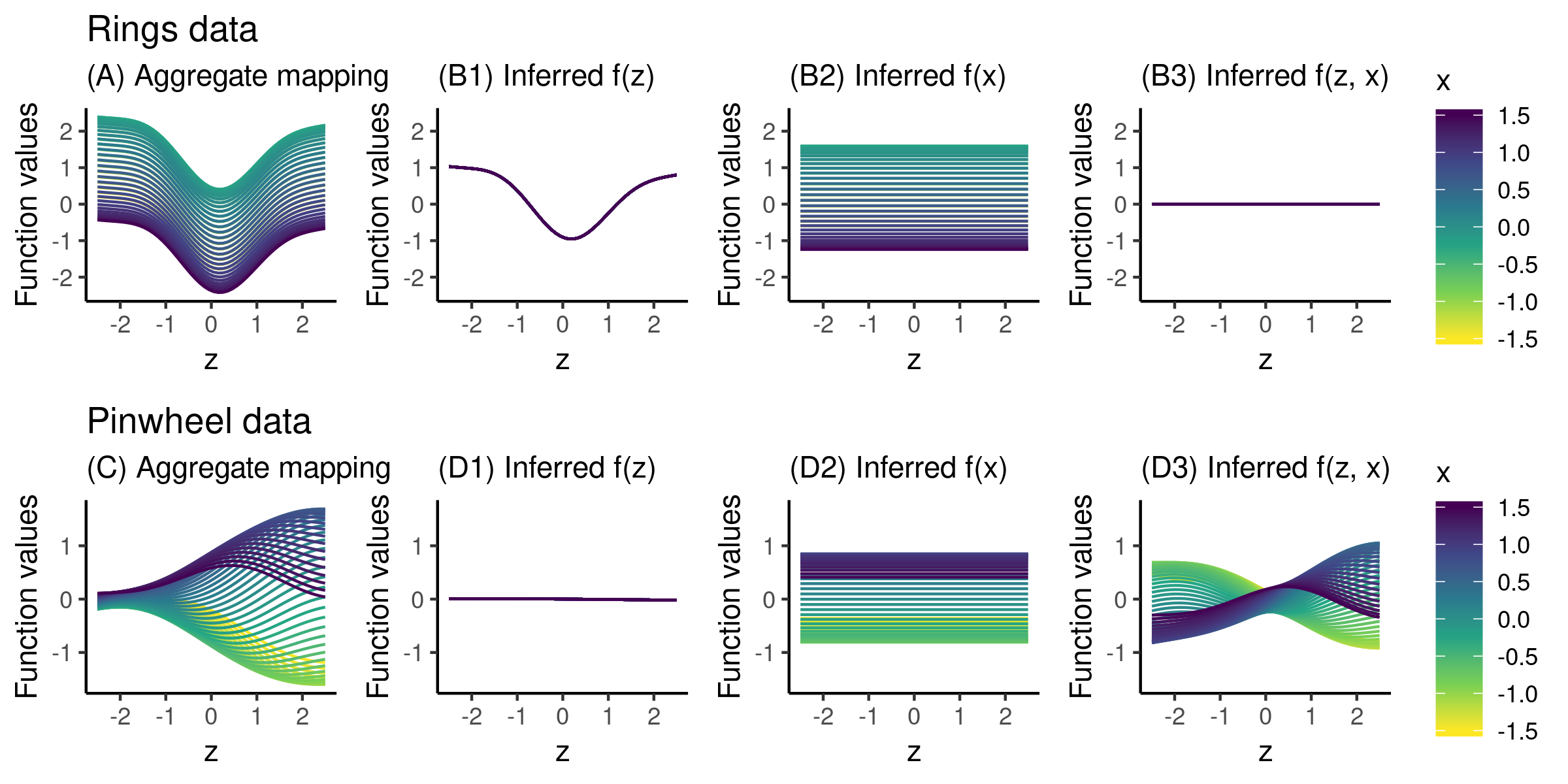}
    \caption{The ADD+INT decomposition inferred by c-GPLVM on the ``rings'' data (top row) and the ``pinwheel data'' (bottom row). The first panels (A and C) show the inferred aggregate function, remaining panels (B1-B3, D1-D3) show the breakdown into three components.}
    \label{fig:circles_and_pinwheel_decomposition}
\end{figure}

First, we consider synthetic two-dimensional datasets as shown as displayed in Figure~\ref{fig:2D_toy_data} (``rings'' and ``pinwheel''). In the observation space, each circle (panel A) or spoke (panel B) corresponds to a \emph{trajectory} and we would like to understand the shared properties of how features vary \emph{along} each trajectory irrespective of the location of each circle or the \emph{angle} of each spoke. If we had additional, \emph{covariate} information on the clustering structure in the form of this angle (shown by colour coding in both panels), we could use this information \emph{jointly} to achieve our goal. Note that the ``rings'' data has an additive data generating mechanism, whereas the ``pinwheel'' features exhibit a complex non-linear interaction between the angle of the spoke and the position along the trajectory. 

The inferred one-dimensional $\boldz$ values are shown by colour coding in the top row of Figures~\ref{fig:circles} and \ref{fig:pinwheel} (panels A1-D1 comparing four models). As expected, the $\boldz$ inferred by GPLVM captures global structure (ignoring covariate information) and $\boldz$ learned by supervised-GPLVM discriminates between the groups defined by the covariate. The Latent Variable Regression, which adjusts for the covariate in a linear model, struggles due to the highly non-linear signal present in the data. 

To gain a better insight into what the models have \emph{learned}, we have also visualised the mean of the posterior predictive distributions in the data space (bottom row of Figures~\ref{fig:circles} and \ref{fig:pinwheel}, same models as in the top row). Here we have varied $\boldz$ from -2 to 2 and kept fixed $\boldx$ to its observed five values. We see that c-GPLVM is the only model which has identified the one-dimensional structure along the circular trajectories for the ``rings'' data and along the spokes for the ``pinwheel'' data. 

Furthermore, we have visualised the ADD+INT decomposition inferred by c-GPLVM. This has been shown for the first coordinate $\boldy_1$ of both data sets in Figure~\ref{fig:circles_and_pinwheel_decomposition}, showing both the inferred aggregate mapping (panels A and C) as well as the individual components (rest of panels). We show here the range of GP mappings obtained by varying $\boldx$ over a fine grid $\boldx \in [-1.5, 1.5]$. 
For the ``rings'' data, c-GPLVM has inferred the presence of additive effects for both $\boldz$ and $\boldx$, but no interaction whereas for the ``pinwheel'' data, all three components contribute. 

The above ``rings'' and ``pinwheel'' exhibit highly non-linear patterns. In Supplementary we explore the behaviour of these models on more realistic synthetic examples, considering dependency structures that we expect to see in real gene expression data.

\subsection{Survival toy example}

\begin{figure}
    \centering
    \includegraphics[width=0.9\columnwidth]{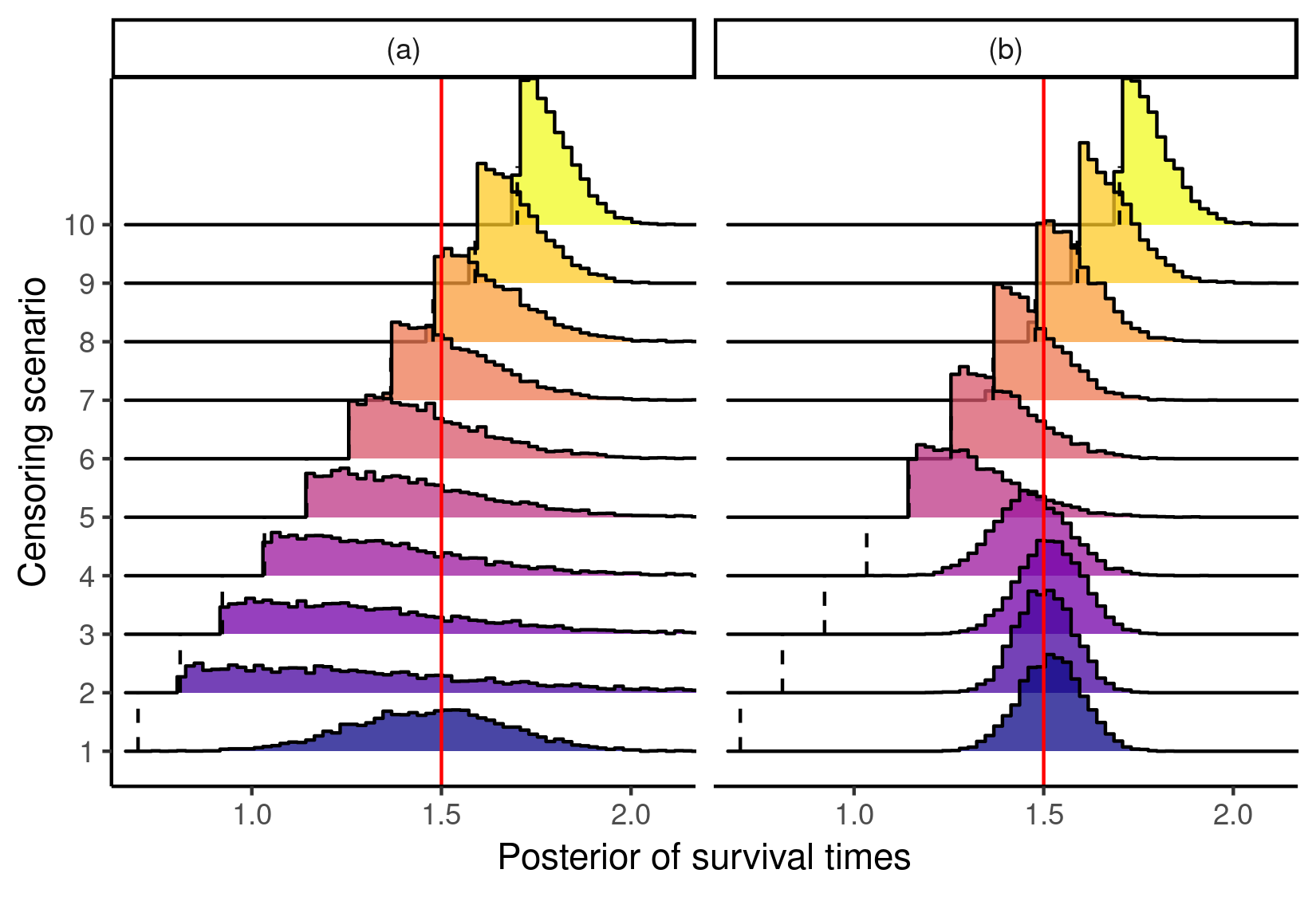}
    \caption{Posterior inference for censored survival times on synthetic data. For two individuals (in panels (a) and (b), demonstrating two different degrees of uncertainty), we vary the lower bound from 0.7 to 1.7 (ten censoring scenarios on $y$-axis), lower bound value denoted by small vertical dashed lines. The true $\boldx$ is always 1.5 (vertical red line).}
    \label{fig:toy_censoring}
\end{figure}

To demonstrate the utility of our inference scheme for censored covariates, we carried out a synthetic experiment. Having generated data as a function of known $\boldz$ and $\boldx$ (details in Supplementary), we wanted to explore the behaviour of our approximate posterior when the true $\boldx = 1.5$, but we vary its lower bound on a grid from 0.7 to 1.7. The resulting approximate posteriors have been shown in Figure~\ref{fig:toy_censoring}. The two panels (a) and (b) represent two observations which exhibit a different degree of uncertainty (in (b) data is more informative of survival than in (a)). Note that when the lower bound becomes larger than the true survival, the posterior starts to concentrate towards the lower bound. 

\subsection{Real-world data experiment}


We now consider a real world data set consisting of $N=770$ breast cancers from The Cancer Genome Atlas cohort \citep{weinstein2013cancer}. We use gene expression measurements across 500 highly variable genes ($\boldY$) and survival information for each patient as a covariate ($\boldx$) to explore the relationship between gene expression changes, survival and a unidimensional latent representation $z$. 

\subsubsection{Prediction of survival times}

To ensure that our modelling assumptions are aligned with the observed data, we first investigate the survival c-GPLVM in a controlled setting. Specifically we focus on its predictive ability. That is, we consider a subset of individuals whose death times have been observed ($N=151$) and carry out artificial censoring in batches of size 5. For 5 patients at a time, we artificially censor their survival time by half a year, and fit c-GPLVM to infer the posterior of the true survivals. When considering alternative methods for the prediction of survival times, we note that most models in the field of survival analysis are non- or semi-parametric, i.e.\ they do not model the baseline hazard function, and thus do not provide a straightforward way for prediction. To obtain a parametric survival model, one common choice is to use the Weibull distribution. Thus, to compare c-GPLVM with a baseline method, we used the Weibull regression model with shrinkage priors as described in \citep{peltola2014hierarchical}, predicting patient survival as a function of the gene expression matrix $\boldY$. 
Predictions by c-GPLVM have been shown against the true values in Figure~\ref{fig:survival_predictions}, detailed comparison with Weibull regression is shown in Supplementary (on repeated sampling from the posterior, the average MSE from the c-GPLVM is 9.2 whereas for the Weibull model it is 17.1). 
We also note that the c-GPLVM \textit{knows when it does not know}, i.e.\ the mis-predictions typically have high uncertainty. 

\begin{figure}
\centering
\includegraphics[width=0.9\columnwidth]{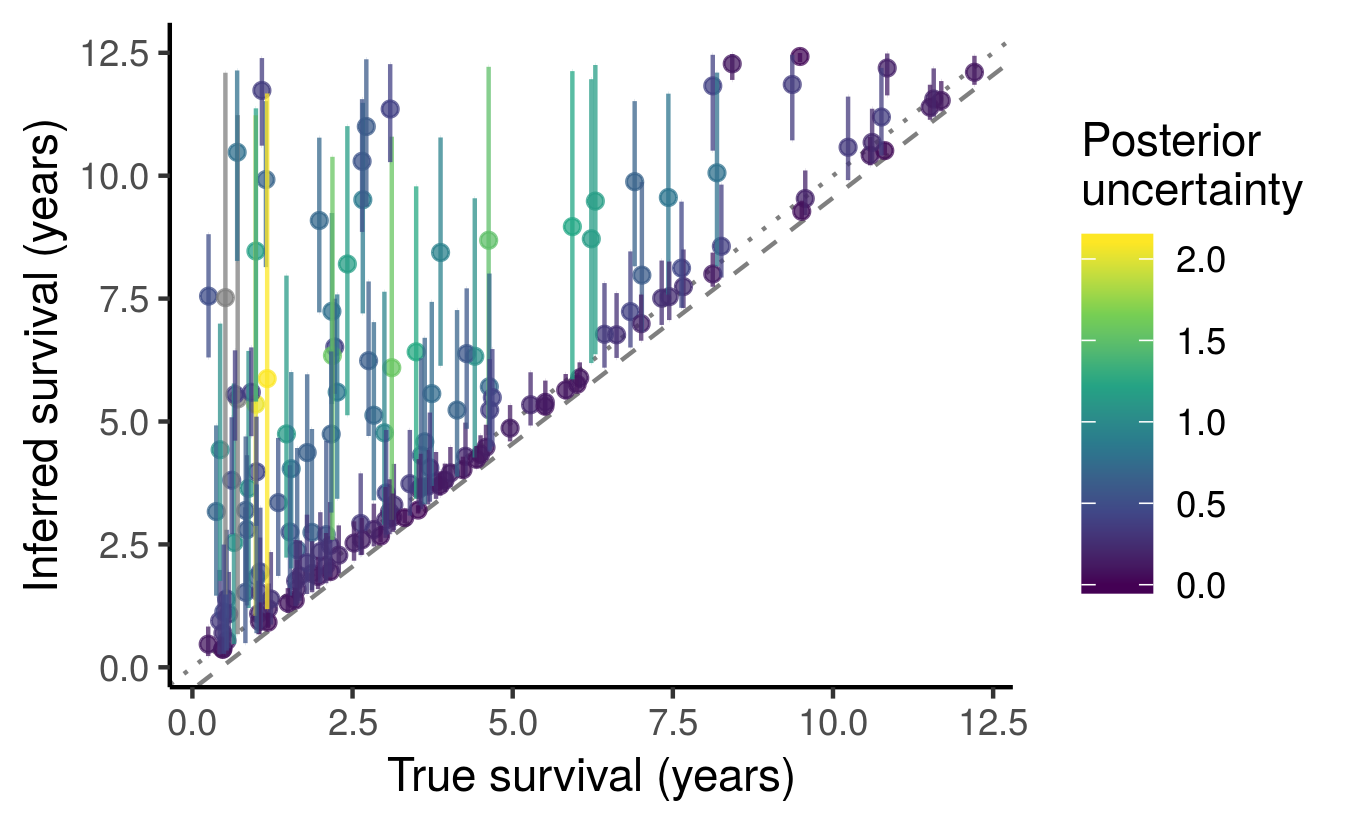}
\caption{Predicted survival times for artificially censored individuals in TCGA data using c-GPLVM (true value on $x$-axis, inferred posterior mean and (5\%, 95\%) quantiles on $y$-axis, colour denotes posterior uncertainty).}
\label{fig:survival_predictions}
\end{figure}

\subsubsection{Survival-adjusted cancer modelling}

\begin{figure}[!t]
\centering
\includegraphics[width=\columnwidth]{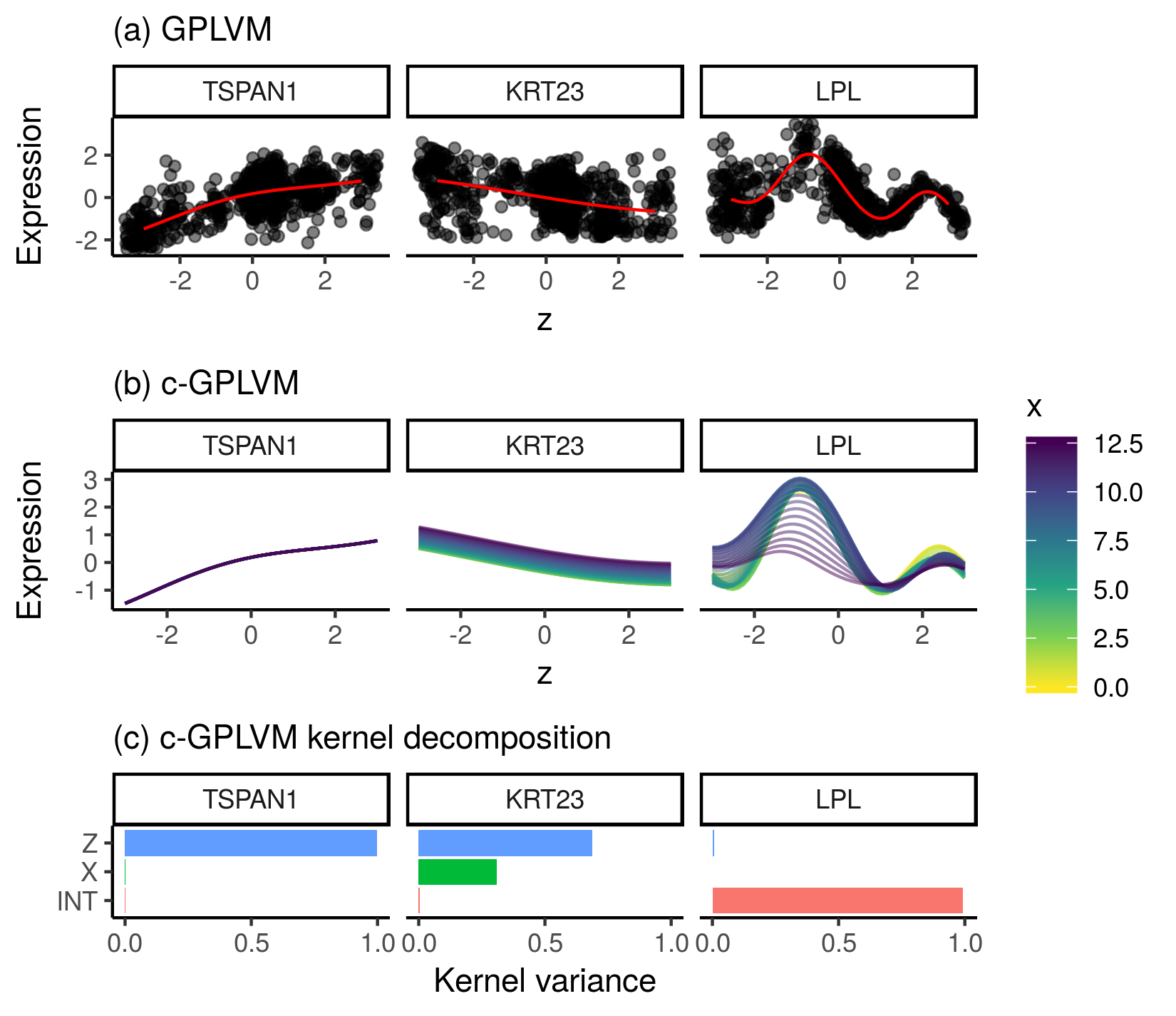}
\caption{GPLVM (panel A) does not capture the variation driven by the censored covariate ($\boldx$=survival), whereas c-GPLVM (panel B) not only captures the effects explained by survival, but in addition, the \addintkernel kernel decomposition (panel C) lets us identify three sets of genes: those without any covariate effect (e.g.\ TSPAN1), genes with an additive covariate effect (e.g.\ KRT23), and genes that exhibit a complex non-linear interaction (e.g.\ LPL).
}
\label{fig:survival_experiment}
\vspace{-5mm}
\end{figure}

Next, we returned to the original data set of  all 770 breast cancers and fitted standard GPLVM and c-GPLVM to the entire cohort. Fig~\ref{fig:survival_experiment}a compares a feature-level fit for standard GPLVM and c-GPLVM for three genes \emph{TSPAN1}, \emph{KRT23} and \emph{LPL}. 
First, \emph{TSPAN1} is a gene that codes for a member of the protein family, Tetraspanins, also known as the transmembrane 4 superfamily. These are small transmembrane glycoproteins which were first described in studies of tumour associated proteins and has been reported to regulate cancer progression in many human cancers \citep{munkley2017cancer}. In line with this, c-GPLVM identifies \emph{TSPAN1} as a gene whose expression increases along the latent coordinate and not with survival suggesting that the identified latent dimension could correspond to disease progression. In contrast, Keratin-23 (\emph{KRT23}) decreases with $z$ but is also additively modulated by the survival covariate with patients who survive longer seemingly retaining a higher expression of \emph{KRT23}. Lipoprotein lipase (LPL) plays a role in breaking down fat in the form of triglycerides, which are carried from various organs to the blood by molecules called lipoproteins. This gene is identified as having interaction effects and we see that there is a range of latent input values $-2 < z < 0$ where low LPL expression is associated with longer survival and high LPL expression is associated with reduced lifespan. This suggests that during certain periods of breast cancer development, high levels of LPL activity could enable aggressive disease progression possibly through acting to provide a supply of fatty acids to fuel tumour growth \citep{kuemmerle2011lipoprotein}. Our analysis here is only illustrative but serves to demonstrate that the kernel decomposition within c-GPLVM allowed us to ``discover" three distinct gene behaviours. This could not have been readily achieved without further post-processing analysis or modification of other existing GPLVM extensions and implementations.

\section{Discussion}

We have introduced the covariate-GPLVM that integrates GP regression and latent variable modelling with a specific focus on the interaction between covariates and latent variables. The c-GPLVM learns latent spaces that  reveal structure in the data which is shared across covariate values. By making use of GP mappings that are specified on the extended joint space of covariates and latent variables, we can model complex non-linear dependencies, while maintaining interpretable and decomposable mappings. The c-GPLVM is applicable for a wide range of applications where there is known structure accounting for variance attributable to this structure. This encompasses scenarios when we are interested in explicitly exploring the interactions between this covariate information and other features, as well as those where such covariates are ``nuisance'' variables and we would like to adjust for confounding factors (e.g.\ batch effects or the presence of different ancestral populations in population genetics). A natural extension of our work is to consider a deep multi-output Gaussian Process formulation \citep{damianou2013deep} in which multiple output dimensions can be coupled via shared Gaussian process mappings. However, this addition introduces many design choices with regards to how this shared structure could be embedded within our structured kernel decomposition framework. We will explore this approach in future iterations of this work.


\subsection*{Acknowledgements} 

We would like to thank Nicolas Durrande for advising us on his prior work on the mean-zero functional constraints. KM is supported by a UK Engineering and Physical Sciences Research Council Doctoral Studentship. KRC is supported by a Banting postdoctoral fellowship from the Canadian Institutes of Health Research, and postdoctoral fellowships from the Canadian Statistical Sciences Institute and the UBC Data Science Institute. CY is supported by a UK Medical Research Council Research Grant (Ref: MR/P02646X/1) and by The Alan Turing Institute under the EPSRC grant EP/N510129/1.

\bibliography{main}
\bibliographystyle{icml2019}

\clearpage

\onecolumn

\begin{appendices}

\section*{Supplementary Information}

\section{ADD+INT kernel decomposition with mean-zero functional constraints}

Suppose $f \sim \GP(0, k(\cdot))$ where $f$ has one-dimensional inputs and $k()$ is the squared exponential kernel, 
\begin{align*}
    k(x, y) = \sigma^2 \exp \left( - \frac{1}{2} \frac{(x-y)^2}{l^2} \right)
\end{align*}

Following \cite{durrande2011additive, durrande2013anova} we can construct a GP prior for $f$ conditional on $\int_a^b f(t) dt = 0$. 
Writing down the joint distribution of $f(x)$ and its integral $\int_a^b f(t) dt$ over some interval $[a, b]$, 

\begin{align*}
\begin{pmatrix}
f(x) \\
\int_a^b f(t) dt
\end{pmatrix} 
&\sim \mathcal{N}
\begin{bmatrix}
    \begin{pmatrix}
    0\\
    0
    \end{pmatrix}\!\!,&
    \begin{pmatrix}
    k(x, x) & \int_a^b k(x, t) dt \\
    \int_a^b k(t, x) dt & \int_a^b \int_a^b k(t, s) dt ds
    \end{pmatrix}
\end{bmatrix}
\end{align*}

we can express the conditional distribution of $f(x)$ conditional on $\int_a^b f(t) dt = 0$, i.e.\ conditional on $f$ being mean-zero. As a result, 
\begin{align*}
    f \; \bigg| \; \left( \int_a^b f(t) dt = 0 \right) \sim \GP(0, \tilde{k}(\cdot)), 
\end{align*}
where
\begin{align*}
    \tilde{k}(x, y) := k(x, y) - 
    \frac{\int_a^b k(x, t) dt \int_a^b k(t, y) dt}{\int_a^b \int_a^b k(t, s) dt ds} \; .
\end{align*}

When $k(x, x')$ is the squared exponential kernel, these integrals have analytic forms, 
\begin{align*}
    \int_a^b k(x, t) dt = 0.5 \sqrt{2\pi} l \sigma^2 
    \left(
        \text{erf}\left( \frac{b-x}{\sqrt{2}l} \right) - \text{erf}\left( \frac{a-x}{\sqrt{2}l} \right)
    \right)
\end{align*}

\begin{align*}
    \int_a^b \int_a^b k(t, s) dt ds = \sqrt{2\pi} l \sigma^2 
    \left(
        (a-b) \text{erf}\left( \frac{a-b}{\sqrt{2}l} \right) + \frac{\sqrt{2}}{\sqrt{\pi}} l (\exp(-\frac{(b-a)^2}{2l^2})-1)
    \right)
\end{align*}

Now, being able to evaluate $\tilde{k}$, we can construct the mean-zero decomposition. We will formulate the ADD+INT kernel decomposition on the joint $(z, x)$ space as follows
\begin{align*}
    k((z, x), (z', x')) := \sigma^2_{b} + \sigma^2_z \tilde{k}_0(z, z') + \sigma^2_x \tilde{k}_0(x, x') + \sigma^2_{zx} \tilde{k}_0(z, z') \tilde{k}_0(x, x')  
\end{align*}
where $\tilde{k}_0$ is the mean-zero kernel as above with kernel variance set to 1.

\section{Additional experiments: Synthetic gene expression data}

The ``rings'' and ``pinwheel'' exhibited highly non-linear patterns, whereas now we explore the behaviour of these models on more realistic synthetic examples, considering dependency structures that we expect to see in real gene expression data. For this purpose, we constructed synthetic data sets where features would exhibit (i) linear additive signal, (ii) linear interactions combined with additive signal, (iii) monotone dependency, and (iv) the latter combined with transient signals. 
In these four scenarios, we measured how accurately do different models uncover the true underlying $\boldz$, results are summarised in Table~\ref{table:corr}. 

\begin{table}[H]
\centering
\caption{Under four data generation schemes (in columns), we compared the four models (in rows) in how accurately they recovered the true $\boldz$, displaying correlation between the true $\boldz$ and the inferred values.}
\small
\begin{tabular}{rcccc}
  \hline
 & linear & linear & monotone & monotone+ \\
 & (ADD) & (INT) &  & transient \\ 
  \hline
 LVR & \textbf{0.99} & \textbf{0.97} & 0.91 & 0.84 \\ 
 GPLVM & 0.81 & 0.82 & 0.80 & 0.91 \\ 
 sup-GPLVM & 0.91 & 0.83 & 0.87 & 0.90 \\ 
 c-GPLVM & \textbf{0.99} & \textbf{0.97} & \textbf{0.96} & \textbf{0.97} \\ 
   \hline
\end{tabular}
\label{table:corr}
\end{table}

In the linear data simulations, the linear assumptions of latent variable regression (LVR) allows it to achieve near perfect recovering of the latent dimension but, despite the increased flexibility, so does c-GPLVM. When the data-generating mechanism is non-linear and including transient effects, the non-linear assumptions of GPLVM and sup-GPLVM exhibit their superiority over LVR but cannot recover the true latent structure as accurately as c-GPLVM. 

\section{Details about the censoring toy example (for section 5.2)}

For the synthetic survival experiment, we generated a four-dimensional data set as follows
\begin{itemize}
    \item $y_i^{(1)} := \sin(z_i) + 0.2x_i + 0.2\sin(z_i) \cdot x_i \cdot I(z_i > 0) + \varepsilon_i$
    \item $y_i^{(2)} := \exp(-z_i^2) + 0.3 \text{tanh}(x_i) + \varepsilon_i$
    \item $y_i^{(3)} := 0.2 z_i + \varepsilon_i$ 
    \item $y_i^{(4)} := \exp(-z_i^2) + \varepsilon_i$
\end{itemize}
where two features depend on $\boldx$ ($\boldy_1$ exhibits an interaction effect and $\boldy_2$ an additive effect), whereas the rest do not. 

Figure~10 illustrates the inferred posteriors for the following two data points (when varying the censoring lower bound):
\begin{itemize}
    \item true $z_i = -2.0$ and $x_i = 1.5$ 
    \item true $z_i = 1.0$ and $x_i = 1.5$ 
\end{itemize}

\section{Details about the artificial censoring experiment on TCGA data (for section 5.3)}

\begin{figure}[H]
    \centering
    \includegraphics[width=\columnwidth]{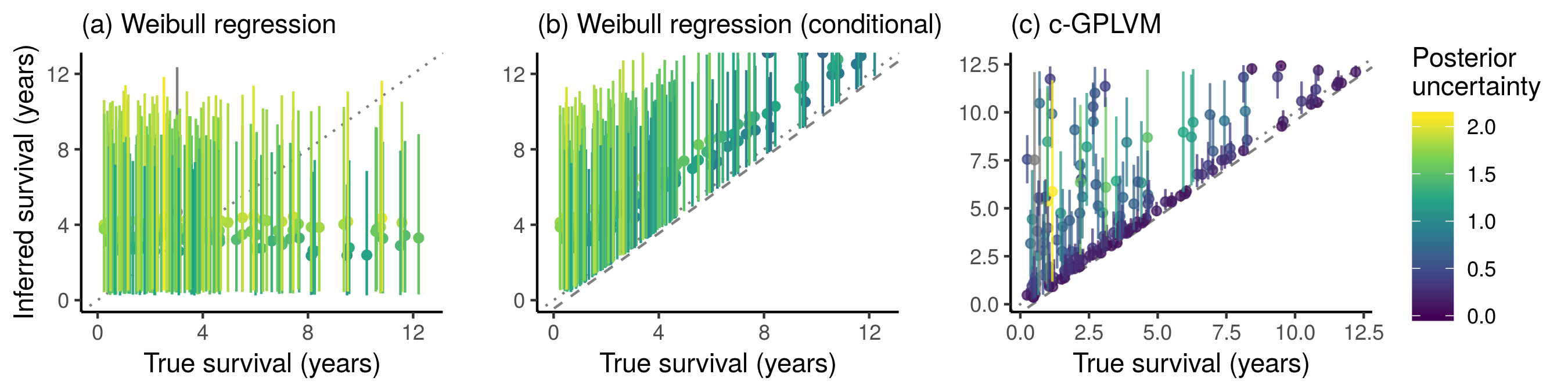}
    \caption{True survival ($x$-axis) and the inferred posterior ($y$-axis) for artificially censored individuals.}
    \label{fig:survival_comparison}
\end{figure}

\begin{figure}[H]
    \centering
    \includegraphics[width=0.4\columnwidth]{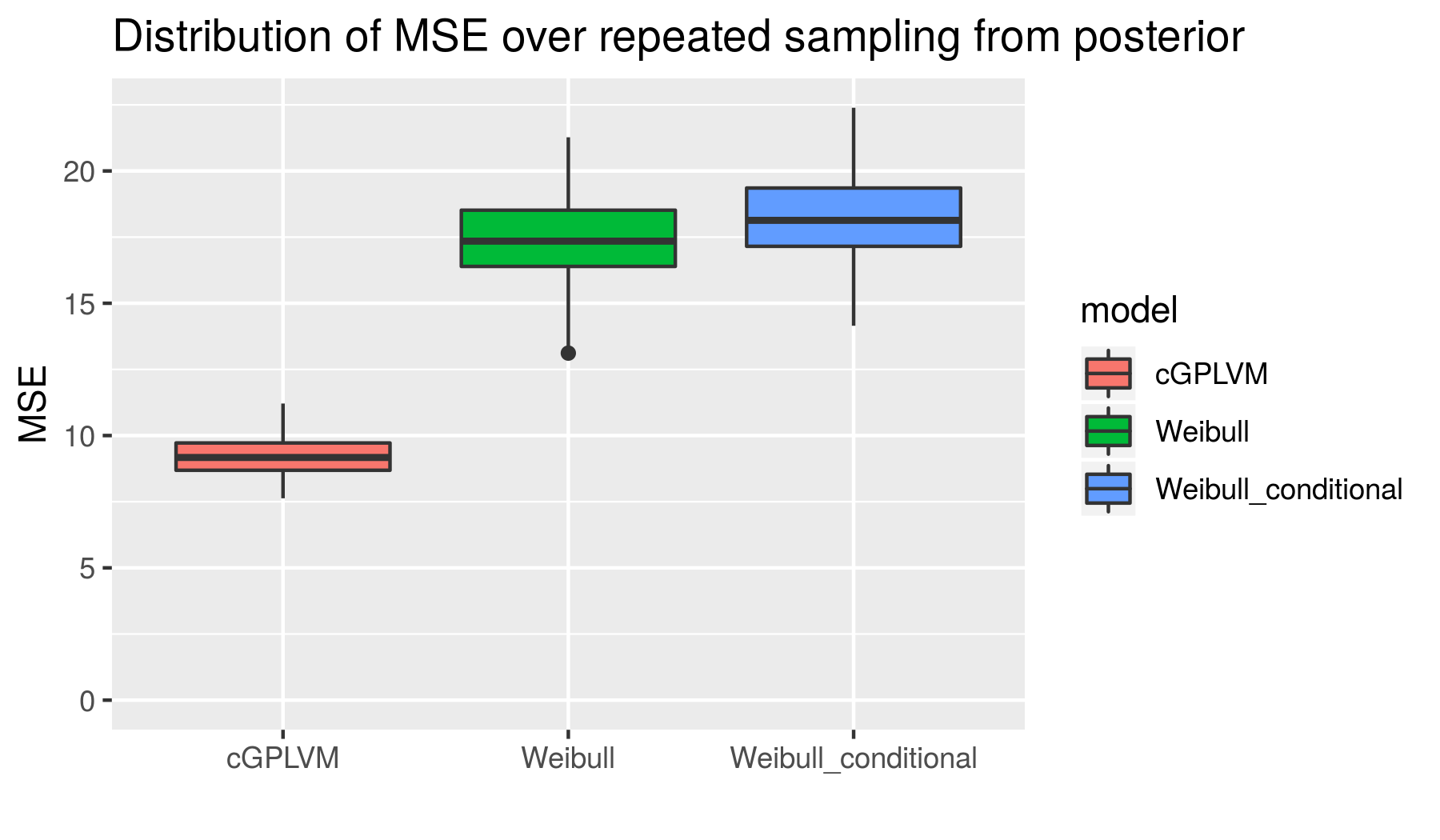}
    \caption{Mean squared prediction error over repeated sampling from the inferred posteriors for the artificial censoring experiment.}
    \label{fig:survival_mse}
\end{figure}

As the posterior predictive from the Weibull regression does not take into account the fact that we have available censoring information, we have provided both the default predictions (which are not informed by censoring times) as well as the \emph{conditional} predictions in Fig~\ref{fig:survival_comparison}. We also quantify the prediction accuracy by repeatedly sampling from the inferred posteriors and show the distribution of mean squared error (MSE) in Figure~\ref{fig:survival_mse}. 

\end{appendices}

\end{document}